# Introducing Dendritic Cells as a Novel Immune-Inspired Algorithm for Anomaly Detection

Julie Greensmith[1], Uwe Aickelin[1], and Steve Cayzer[2]

[1] ASAP, School of Computer Science and IT, University of Nottingham, Jubilee Campus, Wollaton Road, Nottingham, UK, NG8 1BB.
`[jqg,uxa]@cs.nott.ac.uk`

[2] Hewlett-Packard Labs plc, Filton Road, Stoke Gifford, Bristol, UK. BS32 0QZ .
`steve.cayzer@hp.com`

**Abstract.** Dendritic cells are antigen presenting cells that provide a vital link between the innate and adaptive immune system. Research into this family of cells has revealed that they perform the role of co-ordinating T-cell based immune responses, both reactive and for generating tolerance. We have derived an algorithm based on the functionality of these cells, and have used the signals and differentiation pathways to build a control mechanism for an artificial immune system. We present our algorithmic details in addition to some preliminary results, where the algorithm was applied for the purpose of anomaly detection. We hope that this algorithm will eventually become the key component within a large, distributed immune system, based on sound immunological concepts.

## 1 Introduction

In 2003, Aickelin et al outlined a project describing the application of a novel immunological theory, the Danger Theory to intrusion detection systems[1]. The authors of this work suggested that the Danger Theory encompassed pathogenic detection, where the basis for discrimination was not centred around 'self' or 'non-self', but to the presence or absence of danger signals. The paper described how danger signals are released from the body's own tissue cells as a result of necrotic cell death, triggered by an invading pathogen. The immune system was thought to be sensitive to changes in concentration of danger signals and hence an appropriate response is generated. Aickelin et al propose that by differentiating between the chaotic process of necrotic cell death and the safe signals derived from regulated apoptotic cell death, pathogenic agents can be detected within an artificial immune system context.

Currently, the majority of artificial immune systems (AIS) encompass two different types of immune inspired algorithms, namely negative selection (T-cell based), and clonal selection with somatic hypermutation(B-cell based). Exceptions to this include [16], where defined patterns of misbehaviour was used to create danger signals within mobile ad-hoc networks. Danger signals are used in [2] to define the context for collaborative filtering. Implementations including Danger Theory so far, have monitored danger signals directly and have not taken into account any of the cells responsible for signal detection. It is thought that danger signals are detected and processed through 'professional' antigen presenting cells known as dendritic cells. Dendritic cells are viewed as one of the major control mechanisms of the immune system, influencing and orchestrating T-cell responses, in addition to acting as a vital interface between the innate (initial detection) and adaptive (effector response) immune systems.

Dendritic cells (DCs) are responsible for some of the initial pathogenic recognition process, sampling the environment and differentiating depending on the concentration of signals, or perceived misbehaviour, in the host tissue cells. Strong parallels can be drawn from this process to the goal of successful anomaly detection. Current anomaly detection systems frequently rely on profiling 'normal' user behaviour during a training period. Any subsequent observed behaviour that does not match the normal profile (often based on a simple distance metric) is classed as anomalous. At this point an 'alert' is generated. However, these systems can have problems with high levels of false positive errors, as behaviour of users on a system changes over a period of time. Anomaly detection systems remain a high research priority as their inherent properties allow for the detection of novel instances, which could not be detected using a signature based approach. AIS featuring negative selection algorithms have been tried and tested for the purpose of anomaly detection [6]. They produced promising results, but were tarnished by issues surrounding false positives and scalability[8]. Some moderately successful non-AIS systems have been implemented, often involving adaptive sampling[4] and adaptive alert threshold modification.

The aim of this research is to understand the Danger Theory and its implications and to be able to derive an anomaly detection system. More specifically, section 2 of this paper explores the process of cell death and the debate surrounding immune activating signals. Section 3 focuses on dendritic cells with respect to changing morphologies, functions, control of the immune system and in terms of the infectious non-self and danger theories. Section 4 outlines an abstraction from DC functioning and the derivation of a bio-inspired anomaly detection unit. Section 5 shows a worked example of how a DC algorithm can be used as a signal processor, complete with pseudo-code and preliminary results. Section 6 includes a brief analysis of the results and details of future work followed by conclusions.

## 2 Death, Danger and Pathogenic Products

### 2.1 Cell Death & Tissues

Our organs are made up of a collection of specialised cells - generically named tissues. Tissue cells communicate with each other through the use of secreted messenger chemicals known as cytokines. These cytokines can have different effects on the tissue cells in the vicinity and can be either pro or anti-inflammatory in nature. The tissue coupled with the surrounding fluid containing cytokines forms the environment for the DC. The cytokine profile of the tissue changes according to differences in the type of cell death occurring in the tissue at the time, and can be used to assess the state of the tissue.

Pre-programmed cell death, apoptosis is a vital part of the life cycle of a cell. Without it, we would not be able to control the growth of our bodies, and we would be subject to out of control tumours. On the initiation of apoptosis all nuclear material is fragmented in an orderly manner, digestive enzymes are secreted internally and new molecules are expressed on the surface of the cell. The cell is ingested by macrophages, with the membrane still intact. It is thought that the resulting cytokines released from apoptotic cells have an anti-inflammatory effect. However, apoptosis is not the only means by which cells can die. If a cell is subject to stress (by means of irradiation, shock, hypoxia or pathogenic infection), it undergoes the process of necrosis. Due to its unplanned nature, there is no careful repackaging of internal cell contents, or preservation of the membrane. The cell swells up, loses membrane integrity and explodes, releasing its contents into the interstitial fluid surrounding neighbouring tissue cells inclusive of uric acid crystals and heat shock proteins. This type of cytokine environment is said to be pro-inflammatory. This also includes host derived antigens and all other polypeptides which can be phagocytosed by a DC.

The differences in the cytokine profile as a result of cell death are integral for understanding the way in which pathogens and other harmful activities are sensed by the immune system. There have been a number of theories over the last century which have attempted to explain the phenomena of pathogenic recognition. Two of the most hotly debated theories - the Infectious Non-self Model and the Danger Theory are relevant to understanding DCs and imperative to the abstraction of a useful algorithm.

### 2.2 Infectious Non-self - The World According to Janeway

Since 1959 the central tenet of immunology revolved around the specificity of lymphocytes to antigen. According to this theory, proteins belonging to the body (self) are not recognised by the immune system due to the deletion of self reactive T-cells in the thymus. However, this theory did not fit with an amassing volume of evidence. A new perspective emerged in 1989 with Janeway's insightful article [7], which provided an explanation as to why adjuvants added to vaccines were necessary in order to stimulate an immune response. These ideas formed the basis for the infectious non-self model. This model, also known as the detection of microbial non-self, is an augmentation of the long established self non-self principles, though the focus is on innate immune function[5]. This theory proposes that the detection of pathogens is done through the recognition of conserved molecules known as PAMPs (pathogen associated molecular patterns), essentially exogenous signals. PAMPs are produced by all microorganisms irrespective of their pathogenicity, and can be recognised by human immune system cells through the use of pattern recognition receptor e.g. toll-like receptors[13]. The effects of PAMPs on DCs will be explored in more detail in the coming section.

## 2.3 The Danger Theory - The World According to Matzinger

The Danger Theory, proposed by Polly Matzinger in 1994[10], also emphasises the crucial role of the innate immune system for guiding the adaptive immune responses. However, unlike detecting exogenous signals, the Danger Theory rests on the detection of endogenous signals. Endogenous danger signals arise as a result of damage or stress to the tissue cells themselves. The crucial point of the Danger Theory is that the only pathogens detected are the ones that induce necrosis and cause actual damage to the host tissue. The damage can be caused by invading micro-organisms or through defects in the host tissue or innate immune cells. Irrespective of the cause, the danger signals released are always the same. These signals are thought to be derived from the internal contents of the cell[11] inclusive of heat shock proteins, fragmented DNA and uric acid. It is proposed that the exposure of antigen presenting cells to danger signals modulates the cells' behaviour, ultimately leading to the activation of naive T-cells in the lymph nodes. Alternatively, the absence of danger signals and the presence of cytokines released as a result of apoptosis can lead to antigen presentation in a different context, deleting or anergising a matching T-cell[12]. The Danger Theory suggests that the tissue is in control of the immune response.

In [14] it is suggested that DCs have the capability to combine signals from both endogenous and exogenous sources, and respond appropriately. Different combinations of input signals can ultimately lead to the differentiation and activation of T-cells. Both theories have implications for the function of DCs.

## 3 Introducing Dendritic Cells

Dendritic cells (DCs) are white blood cells, which have the capability to act in two different roles - as macrophages in peripheral tissues and organs and as a vehicle for antigen presentation within the secondary lymphoid organs. DCs can be sub-categorised dependent on their location within the body. For the purpose of this investigation and the subsequent algorithm, dermal or tissue resident DCs have been examined. Essentially, the DCs' function is to collect antigen from pathogens and host cells in tissues, and to present multiple antigen samples to naive T-cells in the lymph node. DCs exist in a number of different states of maturity, dependent on the type of environmental signals present in the surrounding fluid. They can exist in either immature, semi-mature or mature forms. The various different phenotypes of DC are shown in Figure 1.

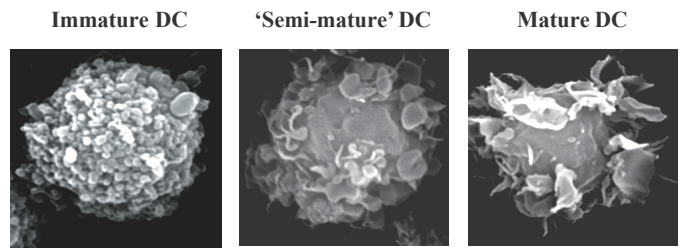

**Immature DC**  **'Semi-mature' DC**  **Mature DC**

Fig. 1. Three differentiation states of DCs as shown from the ESEM photographs shown (see acknowledgements).

### 3.1 Immature DCs

Immature DCs (iDCs) are cells found in their initial maturation state. They reside in the tissue where their primary function is to collect and remove debris from the interstitial fluid. The ingested material is then processed by the cell. It is either metabolised for use by the cell, returned to the environment, or is re-packaged for presentation to another immune cell. At this point the matter can be termed antigen, and could be a 'self' molecule or something foreign. The re-presentation of antigenic material is performed by complexing the antigen with another molecule namely the MHC molecule family, necessary for binding to T-cell receptors. In order to present antigen to T-cells, DC needs sufficient antigen presented with MHC. However, the expression of inflammatory cytokines are needed in order to activate T-cells. Therefore a T-cell encounter with an iDC results in the deactivation of the the T-cell. Differentiation of iDCs occurs in response to the receipt of various signals. This leads to full or partial maturation depending on the combination of signals received.

### 3.2 Mature DCs

Due to the low levels of inflammatory cytokines expressed by iDCs, they are not able to activate T-cells on contact. In order to present antigen and activate T-cells, the increased expression (or up-regulation) of a number of proteins and cytokines is necessary. DCs which have the ability to activate naive T-cells are termed mature DCs (mDCs). For an iDC to differentiate and become a mDC, the iDC has to be exposed to a certain number of signals. This includes activation of toll-like receptors through exposure to both the exogenous and endogenous signals (previously described). On exposure to various combinations of these signals, the DC up-regulates a number of molecules vital for stimulating a T-cell response. Perhaps most importantly, it up-regulates a number of costimulatory molecules, pro-inflammatory cytokines (namely IL-12), and migrates from the tissue to the local draining lymph node. During this migration period, the iDC changes morphologically too. Instead of being compact (optimal for antigen collection), the DC develops whispy, finger-like projections - characterising it as a mDC, as seen in Figure 1. The projections not only make it distinguishable from iDCs, but also increase the surface area of the cell, allowing it to present a greater quantity of antigen.

### 3.3 Semi-Mature DCs

During the antigen collection process, iDCs can experience other environmental conditions. This can affect the end-stage differentiation of a DC. These different conditions can give rise to semi-matureDCs (smDCs). The signals responsible for producing smDCs are also generated by the tissue - endogenous signals. During the process of apoptosis, a number of proteins are actively up-regulated and secreted by the dying cell. The release of TNF-$\alpha$ (tumor necrosis factor) from apoptosing cells is thought to be one candidate responsible for creating semi-mature DCs [9]. As a result of exposure to apoptotic cytokines (TNF-$\alpha$ included), an iDC also undergoes migration to the lymph node, and some maturation as shown in Figure 1. Costimulatory molecules are up-regulated by a small yet significant amount and, after migration to the lymph node, the cell can present antigen to any matching T-cell. However, smDCs do not produce any great amount of pro-inflammatory cytokines, necessary for promoting activation of T-cells. Instead, smDCs can produce small quantities of IL-10 (anti-inflammatory cytokine), which acts to suppress matching T-cells.

### 3.4 Summary

In brief, DCs can perform a number of functions, related to their state of maturation. Modulation between these states is facilitated by the release of endogenous and exogenous signals, produced by pathogens and the tissue itself. The state of maturity of a DC influences the response by T-cells, either immunogenic or tolerogenic, to specific presented antigen. Immature DCs reside in the tissue where they collect antigenic material and are exposed to exogenous and endogenous signals. Based on the combinations of signals, mature or semi-mature DCs are generated. Mature DCs have an activating effect while semi-mature DCs have a suppressive effect. The different cytokine output by the respective cells differ sufficiently to provide the context for antigen presentation. In the following section this information is utilised to derive a signal processor based on the explored functionality of the DCs.

## 4 DC's Meet AIS

There are a number of desirable characteristics exhibited by DCs that we want to incorporate into an algorithm. In order to achieve this, the essential properties, i.e. those that heavily influence immune functions, have to be abstracted from the biological information presented. From this we produce an abstract model of DC interactions and functions, with which we build our algorithm.

### 4.1 Abstraction

As shown, the orchestration of an adaptive immune response via DCs has many subtleties. Only the essential features of this process are mapped in the first instance as we are interested in building an anomaly detector, not an accurate simulation. DCs are examined from a cellular perspective, encompassing behaviour and differentiation of the cells and ignore the interactions on a molecular level and direct interactions with other immune system cells.

DCs have a number of different functional properties that we want to incorporate into an algorithm. Bearing this in mind, we can abstract a number of useful core properties, listed below and represented graphically in Figure 2:

- iDCs have the ability to differentiate in two ways, resulting in mature or semi-mature cells.
- Each iDC can sample multiple antigens within the cell, leading to generalisation of the antigen context.
- The collection of antigen by iDCs is not enough to cause maturity. Exposure to certain signals causes the up-regulation of various molecules that initiate antigen presentation.
- Both smDCs and mDCs show expression of costimulatory molecules, inferring that both types have antigen presenting capabilities.
- The cytokines output by mature and semi-mature cells are different, providing contextual information. The concentration of the output cytokines is dependent on the input signals and can be viewed as an interpretation of the original signal strength.

The effects of individual cytokines and antigen binding affinities have not yet been incorporated into this model, as the initial implementation does not feature T-cells. As stated in [14], we are treating DCs as processors of both exogenous and endogenous signal processors. Input signals are categorised either as PAMPs (P), Safe Signals (S), Danger Signals (D) or Inflammatory Cytokines (IC) and represent a concentration of signal. They are transformed to output concentrations of costimulatory molecules (csm), smDC cytokines (semi) and mDC (mat) cytokines. The signal processing function described in Equation 1 is used with the empirically derived weightings presented in Table 1. These weightings are based on unpublished biological information (see acknowledgements) and represent the ratio of activated DCs in the presence and absence of the various stimuli e.g. approximately double the number of DCs mature on contact with PAMPs as opposed to Danger Signals. Additionally, Safe Signals may reduce the action of PAMPS by the same order of magnitude. Inflammatory cytokines are not sufficient to initiate maturation or presentation but can have an amplifying effect on the other signals present. This function is used to combine each of the input signals to derive values for each of the three output concentrations, where $C_x$ is the input concentration and $W_x$ is the weight.

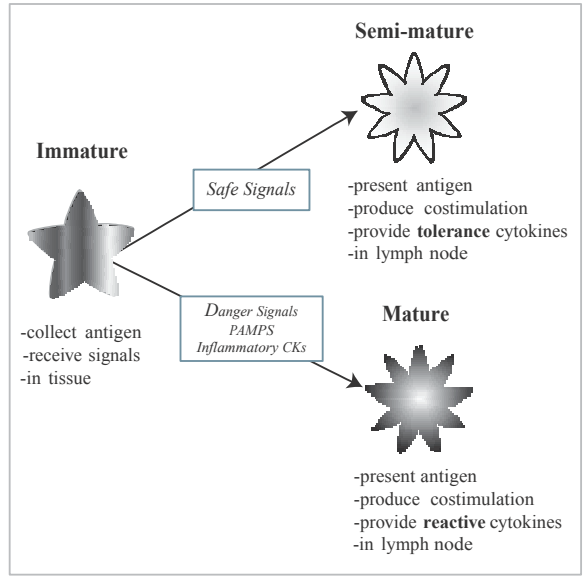

Fig. 2. The iDC, smDC and mDC behaviours and signals required for differentiation. CKs denote cytokines.

$$C_{[csm,semi,mat]} = \frac{(W_P * C_P) + (W_S * C_S) + (W_D * C_D) * (1+IC)}{W_P + W_S + W_D} * 2 \quad (1)$$

Table 1. Suggested weighting values for the signal processing function based on DC maturation ratios

| W | csm | semi | mat |
|---|---|---|---|
| PAMPs(P) | 2 | 0 | 2 |
| Danger Signals(D) | 1 | 0 | 1 |
| Safe Signals (S) | 2 | 3 | -3 |

In order to use this model, input signals have to be pre-classified (either manually or from a signature based intrusion detection system, another anomaly detector, or 'artificial' tissue) based on the following schema:

PAMPs - signals that are known to be pathogenic
Safe Signals - signals that are known to be normal
Danger Signals - signals that may indicate changes in behaviour
Inflammatory Cytokines - signals that amplify the effects of the other signals

In nature, DCs sample multiple antigens within the same section of tissue. To mirror this, we create a population of DCs to collectively form a pool from which a number of DCs are selected for the sampling process, in a similar manner to [17]. An aggregate sampling method should reduce the amount of false positives generated, providing an element of robustness. For such a system to work, a DC can only collect a finite amount of antigen. Hence, an antigen collection threshold must be incorporated so a DC stops collecting antigen and migrates from the sampling pool to a virtual lymph node. In order to achieve this we will use a fuzzy threshold, derived in proportion to the concentration of costimulatory molecules expressed. In order to add a stochastic element, this threshold is within a range of values, so the exact number of antigens sampled per DC varies in line with the biological system.

On migration to the virtual lymph node, the antigens contained within an individual DC are presented with the DC's maturation status. If the concentration of mature cytokines is greater than the semi-mature cytokines, the antigen is presented in a 'mature' context. It is possible to count how many times an antigen had been presented in either context to determine if the antigen is classified as anomalous. In order to crystallise these concepts, a worked example and details of a basic implementation are given in the next section.

## 5  Implementing A DC Based Algorithm

To illustrate the signal processing capabilities of a DC we have designed and implemented a simple prototype system. The purpose of this implementation is to demonstrate the signal processing capability of a population of DCs and their ability to choose between the mature and semi-mature pathways. We expect to see differentiation pathway switching when the data items change from one class to another. In essence a DC algorithm should transform a representation of input data items and signals into the form of antigen-plus-context. From this we can then derive information based on the analysis of the output cytokines.

For such an algorithm to work, some data attributes have to be classed as signals. We use the standard UCI Wisconsin Breast Cancer data-set[15], containing 700 items, each with nine normalised attributes representing the various characteristics of a potentially cancerous cell. Each data item also has a tenth attribute, which is a classification label of class 1 or class 2. Although this is a static dataset, it is suitable for use with our algorithm as data is used in an event driven manner. In order to reduce the difficulty of interpreting the inital experiments only a subset of the data was used. Data items with the largest standard deviation form the danger signals, namely cell size, cell shape, bare nuclei and normal nucleoli. For each of these attributes the mean was calculated over all data items in class 1. Subsequently, the absolute difference from the mean was calculated for each data item, within each attribute, v. The average of the four attribute mean differences comprises the derived danger signal concentration.

To generate concentrations for safe signals and PAMPs, the clump size attribute was chosen as it had the next greatest standard deviation. The median clump size value for all the data items was calculated and each item is compared to the median. If the attribute value is greater than the median, safe signals are derived, equalling the absolute difference between the median and the clump size, and the PAMP concentration is set to zero. If the value is less than the median, then the reverse is true, i.e. safe signals are set to zero and PAMPs are equal to the absolute distance. A worked example is presented in Tables 2 and 3, using one data item and the weightings from Table 1. An example of how to transform the input signals into csms is presented in Equation 2, using a modified version of Equation 1. This example data item was taken from class 1 and, as expected, produces a higher concentration of smDC than mDC cytokines.

Table 2. Sample data item with calculated threshold and signal values (in bold)

| Sample Data Attribute | Data Value | Mean/Threshold | Derived Signal |
|---|---|---|---|
| Clump Size | 10 | 4 | 6 |
| Cell Size | 8 | 6.59 | 1.41 |
| Cell Shape Bare | 8 | 6.56 | 1.44 |
| Nuclei Normal | 4 | 7.62 | 3.62 |
| Nucleoli | 7 | 5.88 | 1.12 |
| Mean Danger Signal | - | - | 1.8975 |

$$C_{csm} = \frac{(2*0)+(2*6)+(1*1.8975)}{2+2+1} \quad (2)$$

Table 3. The output of the signal processing calculations

| Output Signal | Output Conc. |
|---|---|
| csm | 2.7795 |
| semi | 6 |
| mat | -16.1025 |

Although we incorporate inflammatory cytokines into the model, they are not used in this example, as no obvious mapping is available. Antigen is represented in its simplest form, as the identification number of a data item within the dataset. The antigen label facilitates the tracking of data items through the system. Once the signals have been derived and associated with an antigen label, they are processed by the population of DCs. All featured parameters are derived from empirical immunological data. In our experiments, 100 DCs are created for the pool and ten are selected at random to sample each antigen. The signals relating to the antigen are processed by each selected DC and the total amount of output cytokines expressed are measured. The fuzzy migration threshold is set to ten. Once this has been exceeded, a particular DC is removed from the pool and replaced by a new one. After all antigen has been sampled, the context of each antigen is determined based on the number of times it was sampled as either mature or semi-mature. The threshold for classification is derived from the distribution of the data.

The algorithmic details are presented in the pseudo-code as shown in Figure 3:

```
create DC pool of 100 cells

for each data item
    pick 10 DCs from pool
        for each DC
            add antigen(DataLabel) to antigenCollected list
            update input signal concentrations
            calculate concentrations for output cytokines
            update running total of each output cytokine
            if total csms > fuzzy threshold
                removeDC from pool and migrate
                create new DC

for each DC that migrates
    if concentration of semi > mature
        antigenContext = semi
    else
        antigenContext = mature

for each antigen that entered the system
    calculate number of times presented as mature or semi
    if semi > mature
        antigen = benign
    else
        antigen = malignant
```

Fig. 3. Pseudocode for our simple example of a DC algorithm

### 5.1 Experiments and Preliminary Results

Two experiments are performed using the standard Breast Cancer machine learning data-set. This data is divided into class 1 (240 items) and class 2 (460 items). The order of the data items is varied for the two experiments. Experiment 1 uses data on a class by class basis i.e. all of class 1 followed by all of class 2. Experiment 2 uses 120 data items from class 1, all 460 items of class 2 followed by the remaining 120 items from class 1. Each experiment is run 20 times on a Mac iBook G4 1.2MHz, with code implemented in C++(using g++ 3.3). Each run samples each data item 10 times, giving 7000 antigen presentations per run, with 20 runs performed per experiment. The time taken to perform 100 runs is under 60 seconds, giving approximately 10,000 data items sampled per second. The threshold for classification is set to 0.65 to reflect the weighting - items exceeding the threshold are classed as class 2, with lower valued antigen labelled as class 1. These classifications are compared with the labels presented in the original data-set so false positive rates can be measured, in addition to observations of the algorithm's behaviour. Preliminary results are presented in Table 4, and graphically in Figure 4.

Table 4. Table of results to compare two different data orders

| Experiment | Actual Class | Predicted Class 1 | Predicted Class 2 |
|---|---|---|---|
| Experiment 1 | Class 1 | 236 | 4 |
|  | Class 2 | 0 | 460 |
| Experiment 2 | Class 1 | 234 | 6 |
|  | Class 2 | 1 | 459 |

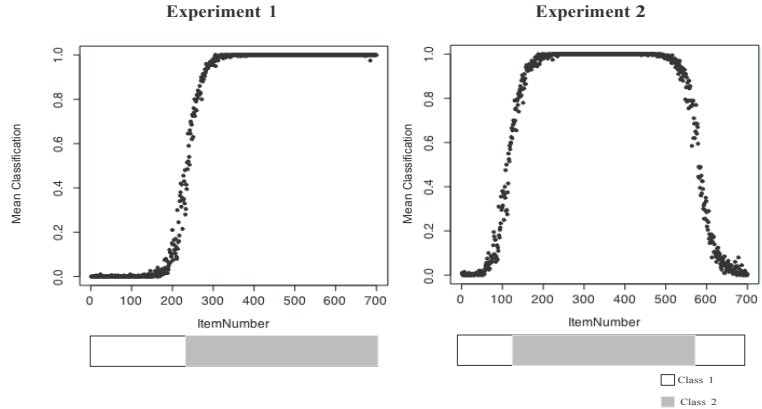

Fig. 4. This figure shows the classification of the 700 items. The bar underneath represents the ordering of the data. The results for the two different data distributions are presented. The y-axis represents the degree of maturity, from 0 (semi-mature, class1) to 1 (mature, class 2) . Data points above the threshold of 0.65 were classified as class 2 and vice-versa.

## 6 Discussion and Future Work

It is important to note that we are not primarily trying to build a new classification algorithm. However, the classification accuracy in these simple experiments exceeds 99%. Rather, we are using this benchmark data-set to show how our dendritic cell model exhibits timely and accurate behavioural switches to changes in context. This is illustrated by our experiments, in which the system rapidly switches to 'danger' mode (Figure 4, Experiment 1) and back again (Figure 4, Experiment 2). Closer examination shows that the misclassifications occur exclusively at the transition boundaries. This is because each DC gathers multiple antigens over a period of time. If an iDC differentiates to an mDC, then every antigen contained in that DC is perceived as dangerous (class 2). Similarly, antigens within an smDC are all perceived as safe (class 1). It is not surprising that during a transition phase there is a small degree of confusion regarding temporally and spatially clustered antigens. A corollary to this is that the DC model is expected to make more mistakes if the context changes multiple times in quick succession; preliminary experiments (results not shown) confirm this. It is important to stress that the data set used was not the ideal application for this algorithm, but it provides data which we can interpret easily to observe the behaviour of the prototype itself.

The implementation of a DC algorithm that we present utilises a relatively simple, well understood data-set. This was useful as it demonstrated the signal processing and change detection potential of a DC based algorithm. However, as stated in the introduction, the ultimate use for this system is as an anomaly detection system with potential applications in computer security. This could be the detection of e-mail worms from an 'outbox'. The presence and type of attachment, rate of sending and content of the mail message could comprise the various signals, with a representation for the content of an attachment and the structure of the message could be an 'antigen'. Alternatively, the algorithm could be used to monitor network behaviour. Various attributes e.g. bandwidth consumption, could be mapped as danger signals, with safe signals and PAMPs derived from the output of various signature matching components e.g. an anti-virus scanner. Antigen could be represented by data flowing through the system in terms of specific patterns of process execution, or perhaps the network packets themselves.

In addition to a more suitable data-set, a number of modifications can be made to the algorithm itself. For instance, we did not include any inflammatory cytokines in our worked example due to data constraints. It would be interesting to explore their proposed amplifying effects on the other signals and on the behaviour across a population of DCs. The current weighting function is simplistic and the weights are empirically derived. Perhaps replacing it with a more sophisticated signal processor based on multi-sensor data fusion techniques would be worth exploring. It will be interesting to see if making the algorithm more biologically plausible results in improved, finer grained detection. Potential improvements could include using a network of cytokines, specifically the cytokines responsible for T-cell activation and proliferation (e.g. IL-12, IL-10, IL-2), and dynamics taken from the accumulating body of immunological experimental results. DCs are only one component of the immune system - the incorporation of other 'cells' such as tissue (for endogenous signals) or T-cells (for an effector response) may give an improved performance.

## 7  Conclusions

In this paper we have presented a detailed description of dendritic cells and the antigen presentation process, from which an algorithm was abstracted. We have also presented a worked example and prototype implementation based on this abstraction The preliminary results are encouraging as both data orders produced low rates of false positive errors.

It is worth making two points about these results. Firstly, it is very encouraging that our simple model illustrates a prediction from the Danger Theory [10]: "...self-reactive killers should be found during the early phases of most responses to foreign antigens, and they should disappear with time". Secondly, it must be remembered that DCs are only part of a system, and that auto-reactive T cells will be tolerised if they subsequently encounter the same antigen in a safe context. A DC model is expected to work in partnership with a T cell system within the larger framework of a distributed immune inspired security system[3].

## 8  Acknowledgements

This project is supported by the EPSRC (GR/S47809/01), Hewlet Packard Labs, Bristol, and the Firestorm intrusion detection system team. ESEM photographs provided courtesy of Dr Julie McLeod, UWE, UK. The authors would like to thank Dr Rachel Harry and Charlotte Williams for the additional biological information. Thanks to Jamie Twycross for assistance throughout. Also, many thanks to Dr Jungwon Kim, William Wilson, Markus Hammonds, Dr Christian Lambert, Gillan Cash, Jim Greensmith and the rest of the Danger Team for their help, support and useful comments.# References

1. U Aickelin, P Bentley, S Cayzer, J Kim, and J McLeod. Danger theory: The link between ais and ids. In Proc. of the Second Internation Conference on Artificial Immune Systems (ICARIS-03), pages 147–155, 2003.
2. Uwe Aickelin and Steve Cayzer. Danger theory and its applications to ais. In Proc. of the Second Internation Conference on Artificial Immune Systems (ICARIS-02), pages 141–148, 2002.
3. Uwe Aickelin, Julie Greensmith, and Jamie Twycross. Immune system approaches to intrusion detection - a review. In Proc. of the Second Internation Conference on Artificial Immune Systems (ICARIS-03), pages 316–329, 2004.
4. E. Eskin, M. Miller, Z. Zhong, G. Yi, W. Lee, and S. Stolfo. Adaptive model generation for intrusion detection. In Proceedings of the ACMCCS Workshop on Intrusion Detection and Prevention, Athens, Greece, 2000., 2000.
5. Ronald N Germain. An innately interesting decade of research in immunology. Nature Medicine, 10(12):1307–1320, 2004.
6. Steven Hofmeyr. An immunological model of distributed detection and its application to computer security. PhD thesis, University Of New Mexico, 1999.
7. Charles A Janeway. Approaching the asymptote? evolution and revolution in immunology. Cold Spring Harb Symp Quant Biol, 1:1–13, 1989.
8. J Kim and P J Bentley. Towards an artificial immune system for network intrusion detection: An investigation of clonal selection with a negative selection operator. In Proceeding of the Congress on Evolutionary Computation (CEC-2001), Seoul, Korea, pages 1244–1252, 2001.
9. Manfred B. Lutz and Gerold Schuler. Immature, semi-mature and fully mature dendritic cells: which signals induce tolerance or immunity? Trends in Immunology, 23(9):991–1045, 2002.
10. P Matzinger. Tolerance, danger and the extended family. Annual Reviews in Immunology, 12:991–1045, 1994.
11. P Matzinger. An innate sense of danger. Seminars in Immunology, 10:399–415, 1998.
12. Polly Matzinger. The danger model: A renewed sense of self. Science, 296:301–304, 2002.
13. Ruslan Medzhitov and Charles A Janeway. Decoding the patterns of self and nonself by the innate immune system. Science, 296:298–300, 2002.
14. Tim R Mosmann and Alexandra M Livingstone. Dendritic cells: the immune information management experts. Nature Immunology, 5(6):564–566, 2004.
15. C.L. Blake S. Hettich and C.J. Merz. UCI repository of machine learning databases, 1998.
16. Slavisa Sarafijanovic and Jean-Yves Le Boudec. An artificial immune system for misbehavior detection in mobile ad-hoc networks with virtual thymus, clustering, danger signal, and memory detectors. In Proc. of the Second Internation Conference on Artificial Immune Systems (ICARIS-04), pages 342–356, 2004.
17. Robert E. Smith, Stephanie Forrest, and Alan S. Perelson. Searching for diverse, cooperative populations with genetic algorithms. Evolutionary Computation, 1(2):127–149, 1993.